\documentclass[runningheads]{llncs}

\usepackage{amsmath}
\usepackage{amsfonts}
\usepackage{graphicx}
\usepackage{microtype}
\usepackage{algorithm}
\usepackage{algpseudocode}
\usepackage{siunitx}
\usepackage{booktabs}

\newcommand{\Req}{\textbf{Require:}\hspace*{0.5em}}

\newcommand{\X}{\hspace*{3mm}}
\newcommand{\XX}{\X\X}
\newcommand{\XXX}{\X\X\X}

\newcommand{\cm}[1]{$\triangleright$ #1}
\newcommand{\loss}{\mathit{loss}}

\title{Deep semi-supervised segmentation with weight-averaged consistency targets}

\author{Christian~S.~Perone\inst{1}\orcidID{0000-0002-1894-6924}, \\	Julien~Cohen-Adad\inst{1,2}\orcidID{0000-0003-3662-9532}}

\authorrunning{C. S. Perone et al.}

\institute{NeuroPoly Lab, Institute of Biomedical Engineering, \\ Polytechnique Montreal, Montreal, QC, Canada.
	\and
	Functional Neuroimaging Unit, CRIUGM, \\ Universite de Montreal, Montreal, QC, Canada.}

\begin{document}
\maketitle

\begin{abstract}
Recently proposed techniques for semi-supervised learning such as Temporal Ensembling and Mean Teacher have achieved state-of-the-art results in many important classification benchmarks. In this work, we expand the Mean Teacher approach to segmentation tasks and show that it can bring important improvements in a realistic small data regime using a publicly available multi-center dataset from the Magnetic Resonance Imaging (MRI) domain. We also devise a method to solve the problems that arise when using traditional data augmentation strategies for segmentation tasks on our new training scheme.
\end{abstract}

\section{Introduction}
In the past few years, we witnessed a large growth in the development of Deep Learning techniques, that surpassed human-level performance on some important tasks \cite{He2015b}, including health domain applications \cite{Rajpurkar2017}. A recent survey \cite{Litjens2017} that examined more than 300 papers using Deep Learning techniques in medical imaging analysis, made it clear that Deep Learning is now pervasive across the entire field. In \cite{Litjens2017}, they also found that Convolutional Neural Networks (CNNs) were more prevalent in the medical imaging analysis, with end-to-end trained CNNs becoming the preferred approach.

It is also evident that Deep Learning poses unique challenges, such as the large amount of data requirement, which can be partially mitigated by using transfer learning \cite{Yosinski2014} or domain adaptation approaches \cite{Ganin2016}, especially in the natural imaging domain. However, in medical imaging domain, not only the image acquisition is expensive but also data annotations, that usually requires a very time-consuming dedication of experts. Besides that, other challenges are still present in the medical imaging field, such as privacy and regulations/ethical concerns, which are also an important factor impacting the data availability.

According to \cite{Litjens2017}, in certain domains, the main challenge is usually not the availability of the image data itself, but the lack of relevant annotations/labeling for these images. Traditionally, systems like Picture Archiving and Communication System (PACS) \cite{Litjens2017}, used in the routine of most western hospitals, store free-text reports, and turning this textual information into accurate or structured labels can be quite challenging. Therefore, the development of techniques that could take advantage of the vast amount of unlabeled data is paramount for advancing the current state of practical applications in medical imaging.

Semi-supervised learning is a class of learning algorithms that can take leverage not only of labeled samples but also from unlabeled samples. Semi-supervised learning is halfway between supervised learning and unsupervised learning \cite{Olivier2006}, where the algorithm uses limited supervision, usually only from a few samples of a dataset together with a larger amount of unlabeled data.

In this work, we propose a simple deep semi-supervised learning approach for segmentation that can be efficiently implemented. Our technique is robust enough to be incorporated in most traditional segmentation architectures since it decouples the semi-supervised training from the architectural choices. We show experimentally on a public Magnetic Resonance Imaging (MRI) dataset that this technique can take advantage of unlabeled data and provide improvements even in a realistic scenario of small data regime, a common reality in medical imaging.

\section{Semi-supervised segmentation using Mean Teacher}
Given that the classification cost for unlabeled samples is undefined in supervised learning, adding unlabeled samples into the training procedure can be quite challenging. Traditionally, there is a dataset $\textbf{X} = (x_i)_{i \in [n]}$ that can be divided into two disjoint sets: the samples $\textbf{X}_l = (x_1, \ldots, x_l)$ that contains the labels $\textbf{Y}_l = (y_1, \dots, y_l)$, and the samples $\textbf{X}_u = (x_{l+1}, \ldots, x_{l+u})$ where the labels are unknown. However, if the knowledge available in $p(x)$ that we can get from the unlabeled data also contains information that is useful for the inference problem of $p(y | x)$, then it is evident that semi-supervised learning can improve upon supervised learning \cite{Olivier2006}. 

Many techniques were developed in the past for semi-supervised learning, usually creating surrogate classes as in \cite{Lee2013}, adding entropy regularization as in \cite{Grandvalet2004} or using Generative Adversarial Networks (GANs) \cite{Salimans2016a}. More recently, other ideas also led to the development of techniques that added perturbations and extra reconstruction costs in the intermediate representations \cite{Rasmus} of the network, yielding excellent results. 
A very successful method called Temporal Ensembling \cite{Laine} was also recently introduced, where the authors explored the idea of a temporal ensembling network for semi-supervised learning where the predictions of multiple previous network evaluations were aggregated using an exponential moving average (EMA) with a penalization term for the predictions that were inconsistent with this target, achieving state-of-the-art results in several semi-supervised learning benchmarks.

In \cite{Tarvainen2017a}, the authors expanded the Temporal Ensembling method by averaging the model weights instead of the label predictions by using Polyak averaging \cite{Polyak1992}. The method described in \cite{Tarvainen2017a} is a student/teacher model, where the student model architecture is replicated into the teacher model, which in turn, get its weights updated as the exponential moving average of the student weights according to:

\begin{align}
	\theta'_t = \alpha \theta'_{t-1} + (1 - \alpha) \theta_{t}
\end{align}

where $\alpha$ is a smoothing hyperparameter, $t$ is the training step and $\theta$ are the model weights. The goal of the student is to learn through a composite loss function with two terms: one for the traditional classification loss and another to enforce the consistency of its predictions with the teacher model.
Both the student and teacher models evaluate the input data by applying noise that can come from Dropout, random affine transformations, added Gaussian noise, among others.

In this work, we extend the mean teacher technique \cite{Tarvainen2017a} to semi-supervised segmentation. To the best of our knowledge, this is the first time that this semi-supervised method was extended for segmentation tasks. Our changes to extend the mean teacher \cite{Tarvainen2017a} technique for segmentation are simple: we use different loss functions both for the task and consistency and also propose a new method for solving the augmentation issues that arises from this technique when used for segmentation. For the consistency loss, we use a pixel-wise binary cross-entropy, formulated as

\begin{equation}
\begin{aligned}
\mathcal{C}(\theta) ={}  \mathbb{E}_{x \in \textbf{X}} \left[ - y \log(p) + (1 - y) \ log(1 - p) \right],
\end{aligned}
\end{equation}


where $p \in [0, 1]$ is the output (after sigmoid activation) of the student model $f(x; \theta)$ and $y \in [0, 1]$ is the output prediction for the same sample from the teacher model $f(x; \theta^\prime)$, where $\theta$ and $\theta^\prime$ are student and teacher model parameters respectively. The consistency loss can be seen as a pixel-wise knowledge distillation \cite{Hinton2015b} from the teacher model. It is important to note that both labeled samples from $\textbf{X}_l$ and unlabeled samples from $\textbf{X}_u$ contribute for the consistency loss $\mathcal{C}(\theta)$ calculation. We used binary cross-entropy, instead of the mean squared error (MSE) used by \cite{Tarvainen2017a} because the binary cross-entropy provided an improved model performance for the segmentation task. We also experimented with confidence thresholding as in \cite{French2017} on the teacher predictions, however, it didn't improve the results. 

\begin{figure}[t]
	\centering
	\includegraphics[width=0.97\linewidth]{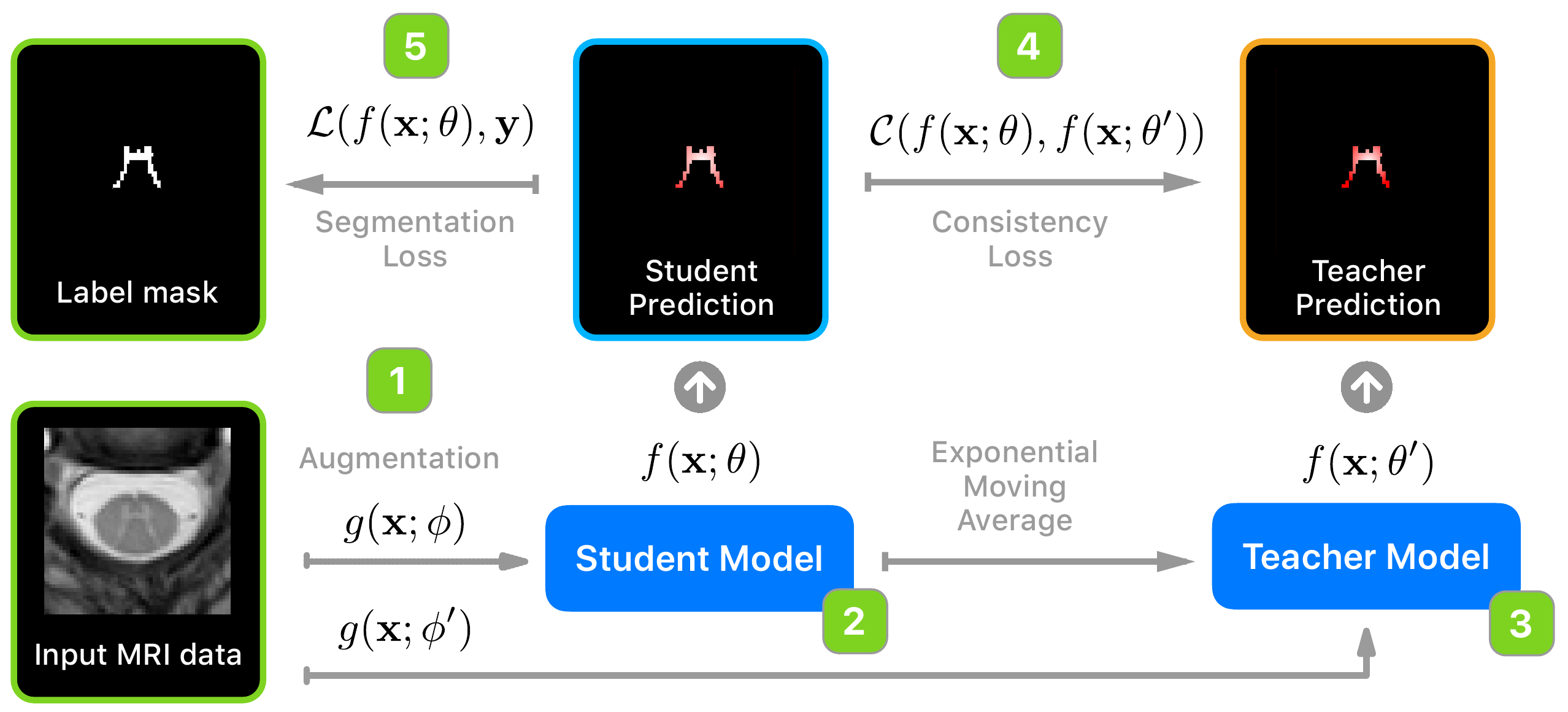}
	\caption{\label{fig:model}
		An overview with the components of the proposed method based on the mean teacher technique. 
		\textbf{(1)}~A data augmentation procedure $g(x; \phi)$ is used to perturb the input data (in our case, a MRI axial slice), where $\phi$ is the data augmentation parameter (i.e. $\mathcal{N}(0,\,\phi$) for a Gaussian noise), note that different augmentation parameters are used for student and teacher models. 
		\textbf{(2)}~The student model.
		\textbf{(3)}~The teacher model that is updated with an exponential moving average (EMA) from the student weights.
		\textbf{(4)}~The consistency loss used to train the student model. This consistency will enforce the consistency between student predictions on both labeled and unlabelled data according to the teacher predictions.
		\textbf{(5)}~The traditional segmentation loss, where the supervision signal is provided to the student model for the labeled samples. 
	}
\end{figure}

For the segmentation task, we employed a surrogate loss for the Dice Similarity Coefficient, called the Dice loss, which is insensitive to imbalance and was also employed by \cite{Perone2017} on the same segmentation task domain we experiment in this paper. The Dice Loss, computed per mini-batch, is formulated as

\begin{align}
L(\theta) = -\frac{2 \sum_i p_i y_i}{\sum_i p_i + \sum_i y_i},
\end{align}

where $p_i \in [0, 1]$ is the $i^{th}$ output (after sigmoid non-linearity) and $y_i \in \{0, 1\}$ is the corresponding ground truth. For the segmentation loss, only labeled samples from $\textbf{X}_l$ contribute for the $\mathcal{L}(\theta)$ calculation. As in \cite{Tarvainen2017a}, the total loss used is the weighted sum of both segmentation and consistency losses.
An overview detailing the components of the method can be seen in the Figure \ref{fig:model}, while a description of the training algorithm is described in the Algorithm \ref{algosemi}.

\begin{algorithm}[h]
	\caption{\label{algosemi}\ 
		Semi-supervised segmentation algorithm. 
	}

	\begin{tabbing}
		\Req $x_i$         \= = training samples \hspace*{25mm} \= \\
		\Req $y_i$         \> = labels for the labeled inputs $i \in \textbf{Y}_l$ \\
		\Req $t$         \> = global step (initialized with zero) \\
		\Req $w(t)$           = consistency weight ramp-up function \\
		\Req $f_\theta(\cdot)$    = neural network model with parameters $\theta$ \\
		\Req $g_\phi(\cdot)$           = stochastic input augmentation procedure with parameters $\phi$ \\
		\X {\bf for} $k$ in $[1,\mathit{num\_epochs}]$ {\bf do} \\
		\XX {\bf for} each minibatch $B$ {\bf do} \\
		\XXX $z_{i \in B} \gets f_\theta(g_\phi(x_{i \in B}))$                              \>\> \cm{evaluate augmented inputs with student model} \\
		\XXX $\tilde{z}_{i \in B} \gets f_{\theta^\prime}(g_{\phi^\prime}(x_{i \in B}))$                      \>\> \cm{teacher model evaluation w/ different perturbations} \\
		\XXX $\loss{}\gets$\=${} \, \mathcal{L}(z, y) + w(t)\frac{1}{|B|}\sum_{i \in B}\mathcal{C}(z_i, \tilde{z}_i)$ \>   \cm{supervised and unsupervised loss components} \\ 		
		\XXX update $\theta$ using, e.g., \textsc{Adam}                                \>\> \cm{update student model parameters} \\
		\XXX $t \gets t + 1$      \> \>                \cm{increment the global step counter} \\
		\XXX $\theta^\prime_t \gets$ \= $\alpha \theta'_{t-1} + (1 - \alpha) \theta_{t}$ \> \cm{update teacher model parameters with using EMA} \\
		\XX {\bf end for} \\
		\X {\bf end for}
	\end{tabbing}

\end{algorithm}

\subsection{Segmentation data augmentation}
In segmentation tasks, data augmentation is very important, especially in the medical imaging domain where data availability is limited,  variability is high and translational equivariance is desirable. Traditional augmentation methods such as affine transformations (rotation, translation, etc) that change the spatial content of the input data, as opposed to pixel-wise additive noise, for example, are also applied with the exact same parameters on the label to spatially align input and ground truth, both subject to a pixel-wise loss. This methodology, however, is unfeasible in the mean teacher training scheme. If two different augmentations (one for the student and another for the teacher) causes spatial misalignment, the spatial content between student and teacher predictions will not match during the pixel-wise consistency loss.

To avoid the misalignment during the consistency loss, such transformations can be applied with the same parametrization both to the student and teacher model inputs. However, this wouldn't take advantage of the stronger invariance to transformations that can be introduced through the consistency loss. For that reason, we developed a solution that applies the transformations in the teacher in a delayed fashion. Our proposed method is based on the application of the same augmentation procedure $g(x; \phi)$ before the model forward pass only for the student model, and then after model forward pass in the teacher model predictions, making thus both prediction maps aligned for the consistency loss evaluation, while still taking leverage of introducing a much stronger invariance to the augmentation between student and teacher models. This is possible because we do backpropagation of the gradients only for the student model parameters.


\section{Experiments}

\subsection{MRI Spinal Cord Gray Matter Segmentation}

In this work, in order to evaluate our technique on a realistic scenario, we use the publicly available multi-center Magnetic Resonance Imaging (MRI) Spinal Cord Gray Matter Segmentation dataset from \cite{Prados2017}.

\subsubsection{Dataset}
 The dataset is comprised of 80 healthy subjects (20 subjects from each center) and obtained using different scanning parameters and also multiple MRI systems. The voxel resolution of the dataset ranges from 0.25x0.25x2.5 mm up to 0.5x0.5x5.0 mm. A sample of one subject axial slice image can be seen in Figure \ref{fig:model}. We split the dataset in a realistic small data regime: only 8 subjects are used as training samples, resulting in 86 axial training slices. We used 8 subjects for validation, resulting in 90 axial slices. For the unlabeled set we used 40 subjects, resulting in 613 axial slices and for the test set we used 12 subjects, resulting in 137 slices. All samples were resampled to a common space of 0.25x0.25 mm.

\subsubsection{Network Architecture}
To evaluate our technique, we used a very simple U-Net \cite{Ronneberger2015} architecture with 15 layers, Batch Normalization, Dropout and ReLU activations. U-Nets are very common in medical imaging domain, hence the architectural choice for the experiment. We also used a 2D slice-wise training procedure with axial slices.

\subsubsection{Training procedure}
For the supervised-only baseline, we used Adam optimizer with $\beta_1 = 0.9$ and $\beta_2 = 0.999$, mini-batch size of 8, dropout rate of 0.5, Batch Normalization momentum of 0.9 and L2 penalty of $\lambda = 0.0008$. For the data augmentation, we used rotation angle between \ang{-4.5} and \ang{4.5} and pixel-wise additive Gaussian noise sampled from $\mathcal{N}(0,\, 0.01)$. We used a learning rate $\eta = 0.0006$ given the small mini-batch size, also subject to a initial ramp-up of 50 epochs and subject to a cosine annealing decay as used by \cite{Tarvainen2017a}. We trained the model for 1600 epochs.

For the semi-supervised experiment, we used the same parameters of the aforementioned supervised-only baseline, except for the L2 penalty of $\lambda = 0.0006$. We used an EMA $\alpha = 0.99$ during the first 50 epochs, later we change it to $\alpha = 0.999$. We also employed a consistency weight factor of 2.9 subject to a ramp-up in the first 100 epochs. We trained the model for 350 epochs.

\subsubsection{Results}
As we can see in Table \ref{tab:results-gray-matter}, our technique not only improved the results on 5/6 evaluated metrics but also reduced the variance, showing a better regularized model in terms of precision/recall balance. The model also showed a very good improvement on overlapping metrics such as Dice and mean intersection over union (mIoU). Given that we exhausted the challenge dataset \cite{Prados2017} to obtain the unlabeled samples, a comparison with \cite{Perone2017} was unfeasible given different dataset splits. We leave this work for  further explorations given that incorporating extra external data would also mix domain adaptation issues into the evaluation.

\begin{table}[]
	\centering
	\caption{Result comparison for the Spinal Cord Gray Matter segmentation challenge using our semi-supervised method and a pure supervised baseline. Results are 10 runs average with standard deviation in parenthesis where bold font represents the best result. Dice is the Dice Similarity Coefficient and mIoU is the mean intersection over union. Other metrics are self-explanatory.}
	\label{tab:results-gray-matter}
\resizebox{\textwidth}{!}{%
	\begin{tabular}{@{}lllllll@{}}
		\toprule
		& \textbf{Dice} & \textbf{mIoU} & \textbf{Accuracy} & \textbf{Precision} & \textbf{Recall} & \textbf{Specificity} \\ \midrule
		Supervised & \begin{tabular}[c]{@{}l@{}}67.915\\ (0.313)\end{tabular} & \begin{tabular}[c]{@{}l@{}}53.679\\ (0.327)\end{tabular} & \begin{tabular}[c]{@{}l@{}}99.745\\ (0.005)\end{tabular} & \begin{tabular}[c]{@{}l@{}}57.948\\ (0.788)\end{tabular} & \textbf{\begin{tabular}[c]{@{}l@{}}92.495\\ (0.907)\end{tabular}} & \begin{tabular}[c]{@{}l@{}}99.775\\ (0.010)\end{tabular} \\ \midrule
		Semi-supervised & \textbf{\begin{tabular}[c]{@{}l@{}}70.209\\ (0.229)\end{tabular}} & \textbf{\begin{tabular}[c]{@{}l@{}}55.509\\ (0.253)\end{tabular}} & \textbf{\begin{tabular}[c]{@{}l@{}}99.792\\ (0.003)\end{tabular}} & \textbf{\begin{tabular}[c]{@{}l@{}}64.732\\ (0.773)\end{tabular}} & \begin{tabular}[c]{@{}l@{}}86.112\\ (0.936)\end{tabular} & \textbf{\begin{tabular}[c]{@{}l@{}}99.846\\ (0.006)\end{tabular}} \\ \bottomrule
	\end{tabular}%
}
\end{table}

\section{Related Work}
Only a few works were developed in the context of semi-supervised segmentation, especially in the field of medical imaging. Only recently, a U-Net was used as auxiliary embedding in \cite{Baur2017}, however, with focus on domain adaptation and using a private dataset.

In \cite{Souly2017a}, they use a Generative Adversarial Networks (GAN) for the semi-supervised segmentation of natural images, however, they employ unrealistic dataset sizes when compared to the medical imaging domain datasets, along with ImageNet pre-trained networks.

In \cite{Xiao2017} they propose a technique using adversarial training, but they focus on the knowledge transfer between natural images with pixel-level annotation and weakly-labeled images with image-level information.

\section{Conclusion}
In this work we extended the semi-supervised mean teacher approach for segmentation tasks, showing that even on a realistic small data regime, this technique can provide major improvements if unlabeled data is available. We also devised a way to maintain the traditional data augmentation procedures while still taking advantage of the teacher/student regularization. The proposed technique can be used with any other Deep Learning architecture since it decouples the semi-supervised training procedure from the architectural choices.

It is evident from these results that future explorations of this technique can improve the results even further, given that even with a small amount of unlabeled samples, we showed that the technique was able to provide significant improvements.

\section{Acknowledgements}
Funded by the Canada Research Chair in Quantitative Magnetic Resonance Imaging (JCA), the Canadian Institute of Health Research [CIHR FDN-143263], the Canada Foundation for Innovation [32454, 34824], the Fonds de Recherche du Québec - Santé [28826], the Fonds de Recherche du Québec - Nature et Technologies [2015-PR-182754], the Natural Sciences and Engineering Research Council of Canada [435897-2013], IVADO, TransMedTech, the Quebec BioImaging Network and NVIDIA Corporation for the donation of a GPU.

\bibliographystyle{splncs}
\bibliography{library}

\end{document}